\documentclass{article} %
\usepackage[final]{colm2025_conference}

\usepackage{microtype}
\usepackage{hyperref}
\usepackage{url}
\usepackage{booktabs}
\usepackage[colorinlistoftodos]{todonotes}
\usepackage{lineno}
\usepackage{amsmath}
\usepackage{subcaption} 
\usepackage{multirow} 
\usepackage{float}
\usepackage{cleveref}

\usepackage{amsmath,amsfonts,bm}

\def\eqref#1{equation~\ref{#1}}

\def\1{\bm{1}}

\DeclareMathAlphabet{\mathsfit}{\encodingdefault}{\sfdefault}{m}{sl}
\SetMathAlphabet{\mathsfit}{bold}{\encodingdefault}{\sfdefault}{bx}{n}

\newtheorem{lemma}{Lemma}[section]

\usepackage[most]{tcolorbox}
\usepackage{xcolor}
\usepackage{enumitem}
\definecolor{correctgreen}{HTML}{2E8B57}
\definecolor{wrongred}{HTML}{C62828}
\definecolor{highlightgray}{gray}{0.95}

\newtcolorbox{answerbox}[2][]{
  colback=highlightgray,
  colframe=black,
  fonttitle=\bfseries,
  title=#2,
  #1
}

\definecolor{darkblue}{rgb}{0, 0, 0.5}
\hypersetup{colorlinks=true, citecolor=darkblue, linkcolor=darkblue, urlcolor=darkblue}

\title{In-context Ranking Preference Optimization}

\usepackage{multirow} 
\usepackage{graphicx}

\author{
Junda Wu$^{1}\thanks{These authors contributed equally to this work.}$,
Rohan Surana$^{1*}$, Zhouhang Xie$^1$, Yiran Shen$^1$, Yu Xia$^1$, \\
\textbf{Tong Yu$^2$, Ryan Rossi$^2$, Prithviraj Ammanabrolu$^1$, Julian McAuley$^1$}
 \\
$^1$UC San Diego \quad 
$^2$Adobe Research \\
\texttt{\{juw069,rsurana,zhx022,jes038,yux078,prithvi,jmcauley\}@ucsd.edu} \\
\texttt{\{tyu,ryrossi\}@adobe.com} \\
}

\definecolor{darkblue}{rgb}{0, 0, 0.5}
\hypersetup{colorlinks=true, citecolor=darkblue, linkcolor=darkblue, urlcolor=darkblue}
\definecolor{myblue}{RGB}{112, 156, 156}
\definecolor{lightblue}{RGB}{237, 250, 252}

\begin{document}

\ifcolmsubmission
\linenumbers
\fi

\maketitle

\begin{abstract}
    Recent developments in Direct Preference Optimization (DPO) allow large language models (LLMs) to function as implicit ranking models by maximizing the margin between preferred and non-preferred responses. In practice, user feedback on such lists typically involves identifying a few relevant items in context rather than providing detailed pairwise comparisons for every possible item pair. Besides, many complex information retrieval tasks, such as conversational agents and summarization systems, critically depend on ranking the highest-quality outputs at the top, further emphasizing the need to support natural and flexible forms of user feedback. To address the challenge of limited and sparse pairwise feedback in the in-context setting, we propose an In-context Ranking Preference Optimization (IRPO) framework that directly optimizes LLMs based on ranking lists constructed during inference. To further capture the natural and flexible forms of feedback, IRPO extends the DPO objective by incorporating both the relevance of items and their positions in the list. Modeling these aspects jointly is non-trivial, as ranking metrics are inherently discrete and non-differentiable, making direct optimization challenging. To overcome this, IRPO introduces a differentiable objective based on positional aggregation of pairwise item preferences, enabling effective gradient-based optimization of discrete ranking metrics. We further provide theoretical insights showing that IRPO (i) automatically emphasizes items with greater disagreement between the model and the reference ranking, and (ii) shows its gradient's linkage to an importance sampling estimator, resulting in an unbiased gradient estimator with reduced variance. Empirical evaluations demonstrate that IRPO outperforms standard DPO approaches in ranking performance, highlighting its effectiveness and efficiency in aligning LLMs with direct in-context ranking preferences.
\end{abstract}

\section{Introduction}

Recent advancements in Direct Preference Optimization (DPO) \citep{rafailov2024direct} 
allow large language models (LLMs) to compare and optimize the pairwise margin \citep{meng2024simpo,wu2024beta} between positive and negative responses without explicit reward functions.
However, in real-world applications (\emph{e.g.}, conversational recommendation \citep{huang2025survey,huang2025towards, surana2025reviews}, generative retrieval \citep{li2025matchinggenerationsurveygenerative}), 
such feedback is typically collected by presenting users with an ordered in-context ranking list and asking them to select relevant items \citep{he2023large,10.1145/3640457.3688191} rather than providing detailed pairwise comparisons,
which highlights the need for frameworks that support natural and flexible feedback formats.
Such in-context feedback on ranked lists yields sparse preference signals that are not directly comparable as explicit pairwise preferences.
In addition, modeling natural and flexible ranking feedback effectively requires capturing both item relevance and positional importance,
of which conventional DPO methods and their underlying preference models (\emph{e.g.}, \citep{rafailov2024direct,chen2024softmax,liu2024lipo}) are limited in modeling directly.
Existing works enable approximations of Plackett-Luce (PL) models for ranking feedback by averaging pairwise Bradley-Terry (BT) comparisons without directly modeling the PL distributions  \citep{zhu2024iterative, chen2024softmax, liu2024lipo}.
Directly applying supervised fine-tuning (SFT) to LLMs is also insufficient for addressing this type of feedback, since ranked-list interactions inherently produce discrete and non-differentiable signals, making direct gradient-based optimization challenging.

To address these limitations of existing approaches, we propose an In-context Ranking Preference Optimization (IRPO) framework 
that integrates design choices across modeling and optimization to better align with real-world ranking feedback. 
IRPO first captures the natural form of user interactions, where users select relevant items from an in-context ranked list without providing exhaustive pairwise comparisons. 
To model such feedback, IRPO employs a PL-inspired positional preference model,
which allows the framework to interpret sparse listwise signals by considering both item relevance and positional importance.
While this formulation improves modeling fidelity, directly optimizing such objectives remains challenging due to the discrete and non-differentiable nature of common ranking metrics. 
To overcome this, IRPO introduces a differentiable objective based on the positional aggregation of pairwise item preferences, enabling effective gradient-based optimization.

To understand the optimization behavior of IRPO, we conduct gradient analysis and provide theoretical insights into its connection to importance sampling gradient estimation. 
Specifically, we show that IRPO acts as an adaptive mechanism that automatically prioritizes items with significant discrepancies between learned and reference policies, 
resulting in efficient and stable optimization.
In addition, we derive an importance-weighted gradient estimator and show that it is unbiased with reduced variance. 
Empirically, we evaluate IRPO across diverse ranking tasks, including conversational recommendation, generative retrieval, and question-answering re-ranking, with various LLM backbones.
Our results consistently show that IRPO significantly improves ranking performance.
We summarize our contributions as follows:
\begin{itemize}
    \item We propose In-context Ranking Preference Optimization (IRPO), a novel framework extending Direct Preference Optimization (DPO) that directly optimizes sparse and in-context ranking feedback.
    \item Specifically we incorporate both graded relevance and positional importance into preference optimization, addressing challenges posed by discrete ranking positions.
    \item We provide theoretical insights linking IRPO's optimization to gradient estimation techniques, demonstrating its computational and analytical advantages.
    \item Extensive empirical evaluations across diverse ranking tasks demonstrate that IRPO achieves consistent performance improvement.
\end{itemize}

\section{Related Work}

\subsection{Ranking Generation with LLMs}
Recent work has leveraged LLMs for ranking across diverse applications—including sequential~\citep{luo2024recranker} and conversational~\citep{yang2024unleashing} recommendation, document retrieval~\citep{liu2024leveraging}, and pair-wise document relevance judgments~\citep{zhuang2023beyond}. Most approaches exploit LLMs’ domain-agnostic strengths rather than improving their intrinsic ranking capabilities~\citep{wu2024surveylargelanguagemodels, pradeep2023rankvicunazeroshotlistwisedocument}, though fine-tuning has been recently explored~\citep{luo2024recranker}. To our knowledge, we are the first to enhance LLM rankings through an alignment framework.
Meanwhile, in contrast to prior works that focus on settings where candidate items are available~\citep{chao-etal-2024-make},
our framework is general and applies to cases where LLMs directly generate a list of responses from input.

\subsection{Direct Preference Optimization for Ranking}
Recent work aligns language models with human feedback via direct preference optimization \citep{rafailov2024direct, meng2024simpo,wu2024beta}. 
Building on learning-to-rank methods \citep{valizadegan2009learning,wu2024coral,wu2021clustering}, 
several approaches recast preference alignment as a ranking task \citep{xie2025survey,zhang2024personalization}. 
For example, GDPO \citep{yao2024no} handles feedback diversity through group-level preferences, and S-DPO \citep{chen2024softmax} extends these ideas to recommendation systems using multiple negatives and partial rankings. However, these approaches generally assume fully supervised or explicitly labeled feedback and require multiple forward passes for optimization, limiting their applicability to more realistic scenarios with sparse and implicit feedback.

Several recent approaches, including Ordinal Preference Optimization (OPO)~\citep{zhao2024ordinal}, Direct Ranking Preference Optimization (DRPO)~\citep{zhou2024optimizing}, and Listwise Preference Optimization (LiPO)~\citep{liu2024lipo}, explicitly leverage differentiable surrogates of ranking metrics such as NDCG to guide optimization. 
OPO uses a NeuralNDCG surrogate built from differentiable sorting (e.g., NeuralSort with Sinkhorn scaling) to align gains with positional discounts, thereby encoding positional importance in a smooth objective. DRPO goes further by explicitly introducing differentiable sorting networks and a margin-based Adaptive Rank Policy Score, and then trains with a differentiable NDCG objective. In contrast, LiPO frames alignment as a listwise learning-to-rank problem and commonly employs \emph{lambda}-weighted pairwise objectives (LiPO-$\lambda$) to approximate listwise metrics like NDCG while operating over all pairs in a list.

In contrast, \textit{IRPO} addresses a fundamentally distinct and practically motivated framework that extends the DPO~\citep{rafailov2024direct} objective by jointly modeling graded relevance and positional importance, and by directly optimizing margins within a single in-context ranking list. Unlike prior works, IRPO does not rely on sorting networks or differentiable sorting approximations, and is designed to eliminate multiple forward passes, enabling a single forward pass per ranked list and lower computational cost.

\section{Preliminaries}

\subsection{Direct Preference Alignment}
DPO \citep{rafailov2024direct} enable reinforcement learning from human feedback (RLHF)~\citep{10.5555/3294996.3295184, 10.5555/3495724.3495977, ouyang2022training} without explicit reward modeling.
Suggested by the Bradley-Terry-Luce (BTL) model of human feedback~\citep{19ff28b9-64f9-3656-ba40-08326a05748e}, 
a response $y_1$ with a reward of $r(x, y_1)$ is preferred over a response $y_2$ with a reward of $r(x, y_2)$ with the probability: 
\begin{equation} \label{eq:btl}
p^*(y_1 \succ y_2 \mid x) = \sigma\left(r(x, y_1) - r(x,y_2)\right),
\end{equation}
where $\sigma(z) = 1/(1+\exp[-z])$ is the sigmoid function.
Suggested in \citep{rafailov2024direct}, the optimal policy in the original RLHF maximization problem with a reference policy \(\pi_{\text{ref}}(y\mid x)\) is given by
\begin{equation}\label{eq:dpo}
    \pi_\theta(y\mid x) = \frac{1}{Z(x)}\,\pi_{\text{ref}}(y\mid x)\exp\!\left(\frac{1}{\beta} r(x,y)\right), 
    Z(x)=\sum_{y}\pi_{\text{ref}}(y\mid x) \cdot \exp\left(\frac{1}{\beta} r(y,x)\right),
\end{equation}
where the partition function $Z(x)$ serves as a normalizer \citep{rafailov2024direct}.
By rearranging \eqref{eq:dpo}, the implicit reward model can be derived and plugged into \eqref{eq:btl},
\begin{equation*}
    r(y,x) = \beta\,\log \frac{\pi^*(y\mid x)}{\pi_{\text{ref}}(y\mid x)} + \beta\,\log Z(x),
\end{equation*}
which formulates the maximum likelihood objective for the target policy $\pi_\theta$ as follows,
\begin{equation*}
\mathcal{L}_\text{DPO}(\pi_\theta; \pi_\text{ref}) = -\mathbb{E}_{(x, y_1, y_2) \sim D} \left[ \log \sigma \left( \beta \log \frac{\pi_\theta(y_1 \mid x)}{\pi_\text{ref}(y_1 \mid x)} - \beta \log \frac{\pi_\theta(y_2 \mid x)}{\pi_\text{ref}(y_2 \mid x)} \right) \right],
\end{equation*}
where $Z(x)$ cancels out and thus directly optimizes the target policy without an explicit reward function.

\subsection{Discounted Cumulative Gain}
Discounted cumulative gain (DCG) has been widely used in various information retrieval and ranking tasks as a metric to capture 
the graded relevance of items in a ranked list while accounting for their positions~\citep{10.1145/3637528.3671687, agarwal2019generalframeworkcounterfactuallearningtorank}. 
Consider a set of candidate items $\boldsymbol{e} = [e_1, e_2, \dots, e_n]$ in a ranking list, where these items represent the potential outputs or responses for a prompt $x$. 
To rank these candidates, \(\tau\) is a permutation over $\{1,2,\dots,n\}$ such that the item at rank $i$ is given by $e_{\tau(i)}$. 
where the ranking vector is defined as $\boldsymbol{k} = [k_1, k_2, \dots, k_n],\: k_i = \tau(i)$, so that $e_{k_i} = e_{\tau(i)}$ is the item placed at the $i$-th position in the ranked list \citep{plackett1975analysis}. 

For each item $e_{k_i}$, users may assign a relevance label $y_{k_i}$ with explicit or implicit feedback \citep{jarvelin2002cumulated}, where its gain is scaled as $G(y_{k_i}) = 2^{y_{k_i}} - 1$. 
In addition, a positional discount factor $d(i) = 1/ \log_2(1+i)$ models the decreasing importance of items placed lower in the ranking list, 
which leads to a weighted gain at position $i$ given as follows,
\begin{equation} \label{eq:gain}
    w(i) = G(y_{k_i}) \cdot d(i) = \frac{2^{y_{k_i}} - 1}{\log_2(1+i)}.
\end{equation}
The quality of a ranked list is measured using the Discounted Cumulative Gain (DCG):
\begin{equation*}
    \text{DCG}(\tau) = \sum_{i=1}^{n} w(i) = \sum_{i=1}^{n} \frac{2^{y_{\tau(i)}} - 1}{\log_2(1+i)}
\end{equation*}
where $w(i)$ represents the gain at rank $i$ for the item with relevance $y_{\tau(i)}$.

\section{IRPO: In-context Ranking Preference Optimization}

\subsection{Preference Modeling}
To capture listwise preferences within the DPO framework, 
for each position $i$ in a ranking list $\tau$, we provide a positional preference model based on pairwise comparisons. 
Following the Plackett-Luce (PL) preference model \citep{plackett1975analysis,luce1959individual}, for the item $e_{\tau(i)}$ at rank $i$,
\begin{equation} \label{eq:pref_model_position}
p^*\Bigl(e_{\tau(i)} \succ \{e_j\}_{j\neq \tau(i)} \mid x\Bigr)
=\sigma\!\Biggl(-\log \sum_{j=1}^{n}\exp\Bigl(s(e_j\mid x)-s(e_{\tau(i)}\mid x)\Bigr)\Biggr),
\end{equation}
where $\sigma(z) = 1/(1+\exp[-z])$ is the sigmoid function and $s(e\mid x)$ is a score function quantifying the quality of item $e$ given the prompt $x$. 
In \eqref{eq:pref_model_position}, we aggregate the in-context pairwise differences between the score of $e_{\tau(i)}$ and all candidate items $e_j$.

To evaluate the entire ranked list, we further aggregate the individual positional preferences into an overall list preference model by weighting each position by its NDCG gain:
\begin{equation}
P^*(\tau\mid x) \propto \prod_{i=1}^{n} \Bigl[ p^*\Bigl(e_{\tau(i)} \succ \{e_j\}_{j\neq \tau(i)} \mid x\Bigr) \Bigr]^{w(i)},
\end{equation}
whose log-likelihood can be derived as follows,
\begin{equation} \label{eq:pref_model_overall}
    \log P^*(\tau\mid x) = \sum_{i=1}^{n} w(i) \cdot \log \sigma\!\Biggl(-\log \sum_{j=1}^{n}\exp\!\Bigl(s(e_j\mid x)-s(e_{\tau(i)}\mid x)\Bigr)\Biggr),
\end{equation}
which ensures that every rank contributes to the assessment of the entire ranking list.

\subsection{Policy Optimization Objective} \label{sec:rdpo}
Following DPO \citep{rafailov2024direct} we plug the reward-based score from \eqref{eq:dpo} into the positional NDCG preference model in \eqref{eq:pref_model_position}, 
to derive our policy optimization objective for IRPO that incorporates graded relevance and the positional importance.
Taking the expectation over the data distribution$(x,\boldsymbol{y})\sim D$ and summing over all ranks, the final IRPO objective is given by:
\begin{align} \label{eq:final_obj}
\mathcal{L}_{\text{IRPO}}(\pi_\theta; \pi_{\text{ref}}) =
-\mathbb{E}_{(x,\boldsymbol{y})}\Biggl[
\sum_{i=1}^{n} w(i) \cdot \log \sigma (z_i)
\Biggr], \quad w(i)=\frac{2^{y_{\tau(i)}}-1}{\log_2(1+i)},
\end{align}
where $w(i)$ is the NDCG gain at rank $i$ according to \eqref{eq:gain}, while the individual rank's preference $z_i$ is 
\begin{equation}\label{eq:margin}
    z_i = -\log \sum_{j=1}^{n} \exp\!\Bigl(\beta\Bigl[\log \frac{\pi_\theta(e_j\mid x)}{\pi_{\text{ref}}(e_j\mid x)} - \log \frac{\pi_\theta(e_{\tau(i)}\mid x)}{\pi_{\text{ref}}(e_{\tau(i)}\mid x)}\Bigr]\Bigr).
\end{equation}
Our formulation naturally extends to other ranking metrics, 
including P@K, MAP, MRR, and eDCG (detailed derivations in Appendix~\ref{app:metrics}), 
by considering their corresponding positional importance,
as a generalized framework for in-context ranking preference optimization.

\subsection{Gradient Analysis and Theoretical Insights}
\label{subsec:4.3}

We further analyze the gradient (detailed derivation in Appendix~\ref{sec:gradient}) by optimizing over IRPO objective to the model parameters $\theta$:
\begin{align} \label{eq:grad}
\nabla_\theta \mathcal{L}_{\text{IRPO}}(\pi_\theta; \pi_{\text{ref}}) = 
\beta\,\mathbb{E}_{(x,\boldsymbol{y})}\left[
\sum_{i=1}^{n} w(i)(1 - \sigma(z_i))
\cdot \sum_{j=1}^n \rho_{ij} \cdot \nabla_\theta \log \frac{\pi_\theta(e_j\mid x)}{\pi_\theta(e_{\tau(i)}\mid x)}
\right],
\end{align}
with the importance weights defined as 
\begin{equation} \label{eq:importance_weights}
\rho_{ij} = \frac{\exp\!\Bigl(\beta\Bigl[\log \frac{\pi_\theta(e_j\mid x)}{\pi_{\text{ref}}(e_j\mid x)} - \log \frac{\pi_\theta(e_{\tau(i)}\mid x)}{\pi_{\text{ref}}(e_{\tau(i)}\mid x)}\Bigr]\Bigr)}
{\sum_{k=1}^{n} \exp\!\Bigl(\beta\Bigl[\log \frac{\pi_\theta(e_k\mid x)}{\pi_{\text{ref}}(e_k\mid x)} - \log \frac{\pi_\theta(e_{\tau(i)}\mid x)}{\pi_{\text{ref}}(e_{\tau(i)}\mid x)}\Bigr]\Bigr)}.
\end{equation}
Following previous DPO methods \citep{rafailov2024direct,chen2024softmax}, 
the gradient term $\nabla_\theta [\log \pi_\theta(e_j\mid x) - \log \pi_\theta(e_{\tau(i)}\mid x)]$ in \eqref{eq:grad}
optimizes the model to further distinguish between the item at rank $i$ and the remaining items in the ranking list.
Intuitively, since $\sum_j \rho_{ij}=1$ for each position $i$, 
the weights $\rho_{ij}$ act automatically as an adaptive mechanism that assigns higher importance to items where the discrepancy between the model and the reference policy is larger. 
In practice, higher importance weights prioritize the gradient contribution from items ranked most wrongly relative to the target ranking.

Inspired by the importance weights $\rho_{ij}$, we further link our gradient analysis to the potential gradient estimator,
\begin{equation}\label{eq:gd-estimator}
    \hat{g}(e_j) = \nabla_\theta \log \frac{\pi_\theta(e_j\mid x)}{\pi_\theta(e_{\tau(i)}\mid x)},
\end{equation}
where the random variable $e_j$ is subordinate to the distribution of importance weights $p(e_j)=\rho_{ij}$ at rank $i$. %
In practice, the gradient calculation could be approximated through importance sampling with the distribution of $p_{i,j}$ at rank $i$. 
We further show the mean and variance properties of the gradient estimator $\hat{g}(e_j)$, 
which serves as an unbiased estimator of the original gradient term and is more efficient in optimization.

\begin{lemma}[Mean Analysis]\label{lem:unbias}
The proposed gradient estimator $\hat{g}(e_j)$ is an unbiased estimation of the gradient term (proof in Appendix~\ref{app:mean}),
\begin{equation}\label{eq:gd}
    g = \sum_{j=1}^n \rho_{ij} \cdot \nabla_\theta \log \frac{\pi_\theta(e_j\mid x)}{\pi_\theta(e_{\tau(i)}\mid x)},
\end{equation}
in \eqref{eq:grad}, which can be achieved by importance sampling with $p(e_j)=\rho_{ij}$ at rank $i$.
\end{lemma}

\begin{lemma}[Variance Analysis]\label{lem:var}
We show that the expected absolute deviation of the proposed estimator is upper bounded as (proof in Appendix~\ref{app:var})
\begin{equation*}
    \mathbb{E}[|\hat{g}(e_j) - g|] 
    \leq \sqrt{\frac{1}{n}\|\rho_{i,j}\|_\infty \cdot \mathbb{E}\left[\|\nabla_\theta \log \frac{\pi_\theta(e_j\mid x)}{\pi_\theta(e_{\tau(i)}\mid x)}\|^2\right]} 
    \leq L\sqrt{\frac{\|\rho_{i,j}\|_\infty}{n}},
\end{equation*}
where $L=\max_{j}{[\nabla_\theta \log \pi_\theta(e_j\mid x) - \nabla_\theta \log \pi_\theta(e_{\tau(i)}\mid x)]}$ and practically clipped to maintain numerical stability \citep{ouyang2022training, schulman2017proximal}.

\end{lemma}

\section{Experiments}

\subsection{Experimental Setup}
\paragraph{Tasks}

To evaluate the effectiveness of IRPO on enhancing LLMs' ranking capabilities, we adopt three tasks: conversational recommendation on Inspired~\citep{hayati-etal-2020-inspired} and Redial~\citep{10.5555/3327546.3327641} dataset; generative (supporting evidence) retrieval on HotpotQA~\citep{yang-etal-2018-hotpotqa} and MuSiQue~\citep{trivedi-etal-2022-musique} dataset; and question-answering as re-ranking on ARC~\citep{clark2018thinksolvedquestionanswering} and CommonsenseQA~\citep{talmor-etal-2019-commonsenseqa} dataset.
Each of the tasks requires LLM to generate a ranked list among a set of candidate answers.
We provide additional detail for task sections in \cref{subsec:conv_rec,subsec:gen_ret,subsec:qa_rerank}.

\paragraph{Baselines}
We compare IRPO against several alignment baselines: Supervised fine-tuning (SFT),
which directly optimizes model outputs from explicit human annotations without preference modeling;
DPO \citep{rafailov2024direct}, which optimizes models from pairwise human preferences by maximizing margins between preferred and non-preferred responses; 
and S-DPO \citep{chen2024softmax}, an extension of DPO tailored for ranking tasks, which leverages multiple negative samples,
inspired by the Plackett-Luce preference model to capture richer ranking signals.
We included more implementation details in Appendix~\ref{app:details}.

\subsection{Conversational Recommendation}
\label{subsec:conv_rec}

\begin{table}[btp]
\centering
\small
\setlength{\tabcolsep}{5pt} %
\begin{tabular}{cc|cccccc|cccccc}
\toprule
 &  & \multicolumn{6}{c|}{Redial} & \multicolumn{6}{c}{Inspired} \\
\multirow{2}{*}{Model} & \multirow{2}{*}{Method} & \multicolumn{3}{c}{NDCG} & \multicolumn{3}{c|}{Recall} & \multicolumn{3}{c}{NDCG} & \multicolumn{3}{c}{Recall} \\
\cmidrule(lr){3-5} \cmidrule(lr){6-8} \cmidrule(lr){9-11} \cmidrule(lr){12-14}
 &  & @1 & @5 & @10 & @1 & @5 & @10 & @1 & @5 & @10 & @1 & @5 & @10 \\
\midrule
\multirow{4}{*}{Llama3} 
 & Base        & 32.5 & 38.2 & 46.3 & 13.0 & 42.6 & 61.2 & 33.8 & 40.8 & 48.7 & 19.8 & 47.7 & 68.1  \\
 & SFT         & 21.6 & 26.5 & 34.9 & 8.1  & 31.2 & 50.4 & 13.8 & 24.7 & 36.2 & 8.1  & 31.7 & 63.3  \\
 & DPO         & 37.6 & 42.5 & 50.1 & 14.9 & 46.5 & 64.0 & 37.5 & 44.0 & \underline{51.2} & 21.9 & 50.2 & 69.2  \\
 & SDPO        & \underline{42.0} & \underline{46.1} & \underline{53.5} & \underline{17.3} & \underline{49.6} & \underline{66.6} & \textbf{53.1} & \textbf{57.6} & \textbf{62.3} & \textbf{30.5} & \underline{63.3} & \underline{75.5}  \\
 & IRPO        & \textbf{74.8} & \textbf{73.6} & \textbf{79.3} & \textbf{27.9} & \textbf{78.1} & \textbf{91.3} & \underline{45.3} & \underline{55.2} & \textbf{62.3} & \underline{22.4} & \textbf{68.8} & \textbf{87.2}  \\
\midrule
\multirow{4}{*}{Phi3} 
 & Base         & 21.1 & 27.4 & 36.8 & 8.7  & 32.1 & 54.0 & 21.1 & 27.4 & 36.8 & 8.7  & 32.1 & 54.0 \\
 & SFT          & 15.9 & 22.2 & 32.7 & 5.8  & 27.2 & 51.6 & 22.1 & 27.0 & 36.6 & 8.5  & 31.2 & 53.4 \\
 & DPO          & \underline{22.5} & \underline{28.9} & \underline{38.2} & \underline{9.2}  & \underline{33.5} & \underline{55.1} & 27.5 & 34.7 & 42.1 & 14.8 & 41.0 & 60.2 \\
 & SDPO         & 21.9 & 23.2 & 28.2 & 8.9  & 23.8 & 32.9 & \underline{32.5} & \underline{36.6} & \underline{44.8} & \textbf{19.6} & \underline{41.2} & \underline{63.3} \\
 & IRPO         & \textbf{35.0} & \textbf{39.3} & \textbf{51.1} & \textbf{12.7} & \textbf{45.6} & \textbf{72.4} & \textbf{42.2} & \textbf{46.3} & \textbf{54.6} & \underline{17.2} & \textbf{54.4} & \textbf{76.0} \\
\midrule
\multirow{4}{*}{Gemma2} 
 & Base         & 19.2 & 26.8 & 38.0 & 7.2  & 32.8 & 58.5 & 15.3 & 22.4 & 31.7 & 7.7  & 28.5 & 52.0 \\
 & SFT          & 25.6 & 30.2 & 39.3 & 9.6  & 34.4 & 55.6 & 15.0 & 25.9 & 35.5 & 7.5  & 32.9 & 57.9 \\
 & DPO          & 30.1 & 36.0 & 44.3 & 12.1 & 40.4 & 59.4 & 27.5 & 34.7 & 42.1 & 14.8 & 41.0 & \underline{60.2} \\
 & SDPO         & \underline{48.5} & \underline{52.7} & \underline{59.2} & \underline{19.6} & \underline{55.8} & \underline{70.9} & \underline{40.0} & \underline{44.0} & \underline{48.4} & \textbf{22.7} & \underline{48.2} & 60.0 \\
 & IRPO         & \textbf{68.8} & \textbf{71.4} & \textbf{77.3} & \textbf{25.1} & \textbf{77.0} & \textbf{90.5} & \textbf{42.2} & \textbf{46.3} & \textbf{54.6} & \underline{17.2} & \textbf{54.4} & \textbf{76.0} \\
\bottomrule
\end{tabular}
\caption{
Performance on Redial and Inspired of Conversational Recommendation, evaluated using NDCG and Recall for the top 1, 5, and 10 predictions out of 20 candidate items.
}
\label{tab:main-crs}
\vspace{-1em}
\end{table}

For conversational recommendation, we use two widely adopted datasets,  
Inspired~\citep{hayati-etal-2020-inspired} and Redial~\citep{10.5555/3327546.3327641}.
Following~\citep{he2023large,10.1145/3640457.3688191,10.1145/3700604,jiang2024utilityevaluatingllmrecommender},
LLMs generate ranked lists of $20$ candidate movies per dialogue context.
To evaluate IRPO and baselines in generating ranked conversational recommendation lists, 
we follow~\citep{he2023large} and construct candidate movie sets for each dialogue context (detailed in~\Cref{app:details}).
These candidate movies are then assigned relevance scores reflecting their importance in calculating the NDCG gain for IRPO in \eqref{eq:final_obj}, 
where ground-truth movies receive a score of 2, GPT-generated movies a score of 1, and random movies a score of 0.
Following~\citep{he2023large,10.1145/3640457.3688191,10.1145/3700604,jiang2024utilityevaluatingllmrecommender}, 
we report the Recall and NDCG at top-$k$ positions.

Results in~\Cref{tab:main-crs} show that supervised fine-tuning (SFT) can be harmful for most metrics and datasets compared with base model performance, 
due to strong popularity bias in conversational recommendation datasets, which is likely to cause model overfitting \citep{gao2025processsupervisedllmrecommendersflowguided,10.1145/3511808.3557423,10.1007/s11257-024-09406-0}.
With multi-negative policy optimization (SDPO), such biasing effects could be alleviated, leading to relatively better performance compared to DPO and SFT.
In addition, we show that IRPO achieves consistently better or comparable performance across datasets in conversational recommendation compared with baselines,
by further considering positional importance for each item, weighted by NDCG weights based on the relevancy score feedback from users.
With the complete ranking list optimized by pairwise comparative margins measured by DPO~\citep{rafailov2024direct},
IRPO acts automatically as an adaptive mechanism that assigns higher importance to items where the discrepancy between the model and the reference policy is larger.

\begin{table}[btp]
\centering
\small
\setlength{\tabcolsep}{5pt} %
\begin{tabular}{cc|cccccc|cccccc}
\toprule
 &  & \multicolumn{6}{c|}{HotpotQA} & \multicolumn{6}{c}{MuSiQue} \\
\multirow{2}{*}{Model} & \multirow{2}{*}{Method} & \multicolumn{3}{c}{NDCG} & \multicolumn{3}{c|}{Recall} & \multicolumn{3}{c}{NDCG} & \multicolumn{3}{c}{Recall} \\
\cmidrule(lr){3-5} \cmidrule(lr){6-8} \cmidrule(lr){9-11} \cmidrule(lr){12-14}
 &  & @1 & @3 & @5 & @1 & @3 & @5 & @1 & @3 & @5 & @1 & @3 & @5 \\
\midrule
\multirow{4}{*}{Llama3} 
 & Base   & 21.8 & 31.3 & 38.2 & 19.1 & 38.4 & 54.4 & 42.5 & 43.5 & 51.9 & 18.8 & 44.9 & 60.7 \\
 & SFT    & 35.3 & 41.4 & 48.3 & 29.5 & 46.3 & 62.1 & 36.9 & 39.1 & 48.0 & 16.6 & 40.7 & 57.6 \\
 & DPO    & 31.2 & 39.5 & 46.3 & 27.3 & 45.6 & 61.0 & 51.1 & 53.9 & 60.8 & 21.6 & 55.8 & 68.9 \\
 & SDPO   & \underline{41.3} & \underline{48.9} & \underline{54.6} & \underline{36.2} & \underline{54.7} & \underline{68.1} & \underline{58.6} & \textbf{60.8} & \underline{66.7} & \underline{26.0} & \textbf{62.3} & \underline{73.3} \\
 & IRPO   & \textbf{94.6} & \textbf{97.2} & \textbf{97.5} & \textbf{83.6} & \textbf{99.1} & \textbf{99.8} & \textbf{65.1} & \underline{59.4} & \textbf{69.4} & \textbf{27.6} & \underline{59.2} & \textbf{78.0} \\
\midrule
\multirow{4}{*}{Phi3} 
 & Base & 24.0 & 32.1 & 39.4 & 21.2 & 38.3 & 55.8 & 23.9 & 24.8 & 31.8 & 9.8  & 26.0 & 39.2 \\
 & SFT  & 26.0 & 34.1 & 41.1 & 23.2 & 40.2 & 56.3 & 33.0 & 32.2 & 41.3 & 14.1 & 33.4 & 50.5 \\
 & DPO  & \underline{45.8} & \underline{52.5} & \underline{57.6} & \underline{40.6} & \underline{57.7} & \underline{69.5} & 31.6 & \underline{34.4} & 43.5 & 13.5 & 36.5 & 53.7 \\
 & SDPO & 45.5 & 52.3 & 57.4 & 40.2 & 57.4 & 69.3 & \textbf{41.9} & \textbf{43.9} & \underline{52.0} & \underline{18.4} & \underline{45.5} & \underline{60.7} \\
 & IRPO & \textbf{61.2} & \textbf{73.8} & \textbf{78.5} & \textbf{53.8} & \textbf{82.9} & \textbf{93.7} & \underline{41.5} & \textbf{43.9} & \textbf{52.4} & \textbf{18.5} & \textbf{45.7} & \textbf{61.7} \\
\midrule
\multirow{4}{*}{Gemma2} 
 & Base & 32.5 & 39.6 & 45.4 & 28.5 & 45.1 & 58.4 & 40.9 & 42.8 & 51.1 & 17.9 & 44.3 & 59.9 \\
 & SFT  & 19.7 & 27.8 & 35.4 & 16.6 & 33.7 & 51.1 & 39.8 & \underline{43.3} & \underline{51.7} & 17.2 & \underline{45.3} & \underline{61.3} \\
 & DPO  & \underline{69.1} & \underline{72.6} & \underline{75.5} & \underline{61.5} & \underline{75.3} & \underline{81.9} & 40.9 & 43.1 & 51.1 & \underline{18.1} & 44.9 & 59.8 \\
 & SDPO & 53.3 & 59.0 & 63.2 & 47.7 & 62.9 & 72.5 & 40.6 & 42.3 & 50.6 & 17.9 & 43.7 & 59.4  \\
 & IRPO & \textbf{94.5} & \textbf{96.8} & \textbf{97.4} & \textbf{83.5} & \textbf{98.5} & \textbf{99.8} & \textbf{55.4} & \textbf{51.0} & \textbf{61.1} & \textbf{23.8} & \textbf{51.6} & \textbf{70.6} \\
\bottomrule
\end{tabular}
\caption{
Performance on HotpotQA and MuSiQue for Generative Retrieval, evaluated using NDCG and Recall for the top 1, 3, and 5 predictions out of 10 candidate contexts.
}
\label{tab:main-genrag}
\vspace{-1em}
\end{table}

\subsection{Generative Retrieval}
\label{subsec:gen_ret}
For generative retrieval, we evaluate IRPO using multi-hop question-answering datasets, 
HotpotQA~\citep{yang-etal-2018-hotpotqa} and MuSiQue~\citep{trivedi-etal-2022-musique} (detailed in~\Cref{app:details}). 
Following prior work~\citep{shen-etal-2024-retrieval,xia2025knowledgeawarequeryexpansionlarge}, 
we prompt LLMs to rank a set of candidate context paragraphs per question. 
We assign binary relevance scores, setting a score of 1 for supporting contexts and 0 for distractors.

Based on the comparative results in~\Cref{tab:main-genrag}, SFT consistently reduces retrieval effectiveness relative to base models. 
This occurs because SFT tends to over-optimize policies, limiting their ability to generalize effectively to the nuanced retrieval challenges inherent in multi-hop queries.
While SDPO, leveraging multi-negative sampling, occasionally achieves better top-1 performance compared with IRPO, 
IRPO attains substantial improvements in NDCG and Recall across the entire ranking list, 
demonstrating its effectiveness in explicitly modeling positional importance.
By optimizing pairwise comparative margins comprehensively across entire ranking lists,
IRPO adaptively prioritizes contexts with larger divergences between model predictions and preference feedback. 
Thus, IRPO offers robust generalizability and better retrieval accuracy for complex generative retrieval tasks.

\subsection{Question-answering as Re-ranking}
\label{subsec:qa_rerank}
In the question-answering as re-ranking scenario, we assess the ability of LLMs to identify correct answers from multiple choices based on contextual relevance. 
We evaluate IRPO on two widely-used datasets, ARC~\citep{clark2018thinksolvedquestionanswering} and CommonsenseQA~\citep{talmor-etal-2019-commonsenseqa}  (detailed in~\Cref{app:details}).
Each candidate is explicitly assigned binary relevance: the correct answer receives a score of 1, 
and incorrect answers a score of 0.

Comparative results summarized in~\Cref{tab:main-qa} highlight IRPO’s robust improvements across models and datasets.
While SFT achieves relatively better performance compared to its performance on other tasks, 
it still struggles to prioritize and disambiguate the correct answer among similar candidate answers.
On the other hand, SDPO, benefiting from multi-negative optimization, 
typically yields better performance by distinguishing subtle semantic differences among distractors.
IRPO further demonstrates superior effectiveness in this challenging ranking task, 
substantially outperforming all baseline approaches.
Aligned with our theoretical insights into IRPO optimization in \Cref{subsec:4.3}, IRPO boosts performance by adaptively focusing on candidates with greater discrepancies, 
enabling comprehensive comparisons that effectively resolve subtle semantic differences and yield substantial gains on challenging datasets like ARC and CommonsenseQA.

\begin{table}[btp]
\centering
\small
\setlength{\tabcolsep}{5pt} %
\begin{tabular}{cc|cccccc|cccccc}
\toprule
 & & \multicolumn{6}{c|}{ARC}  & \multicolumn{6}{c}{CommonsenseQA} \\
\cmidrule(lr){3-8} \cmidrule(lr){9-14}
\multirow{2}{*}{Model} & \multirow{2}{*}{Method} & \multicolumn{3}{c}{NDCG} & \multicolumn{3}{c|}{Recall} & \multicolumn{3}{c}{NDCG} & \multicolumn{3}{c}{Recall} \\
\cmidrule(lr){3-5} \cmidrule(lr){6-8} \cmidrule(lr){9-11} \cmidrule(lr){12-14}
 &  & @1 & @3 & @5 & @1 & @3 & @5 & @1 & @3 & @5 & @1 & @3 & @5 \\
\midrule
\multirow{4}{*}{Llama3} 
 & Base & 13.2 & 21.7 & 28.6 & 13.2 & 28.4 & 44.9 & 24.1 & 32.6 & 38.9 & 24.1 & 39.3 & 54.8 \\
 & SFT  & 13.5 & 21.4 & 28.4 & 13.5 & 27.7 & 44.6 & 22.8 & 31.6 & 38.7 & 22.8 & 38.6 & 56.1 \\
 & DPO  & 15.2 & 22.8 & 30.4 & 15.2 & 29.1 & 47.3 & \underline{54.4} & 59.5 & 63.4 & \underline{54.4} & 63.6 & 73.2 \\
 & SDPO & \underline{23.4} & \underline{28.3} & \underline{35.4} & \underline{23.4} & \underline{32.5} & \underline{50.0} & 54.2 & \underline{59.9} & \underline{63.9} & 54.2 & \underline{64.4} & \underline{74.2} \\
 & IRPO & \textbf{27.4} & \textbf{38.9} & \textbf{46.5} & \textbf{27.4} & \textbf{47.6} & \textbf{66.3} & \textbf{68.6} & \textbf{83.1} & \textbf{85.0} & \textbf{68.6} & \textbf{92.8} & \textbf{97.5} \\
\midrule
\multirow{4}{*}{Phi3} 
 & Base & \underline{9.8} & 17.9 & 25.1 & \underline{9.8} & 24.3 & 41.9 & 24.3 & 33.2 & 38.7 & 24.3 & 40.0 & 53.6  \\
 & SFT  & 8.8 & 17.7 & \underline{25.7} & 8.8 & \underline{24.7} & \textbf{44.3} & 24.4 & 32.6 & 39.6 & 24.4 & 39.0 & 56.3  \\
 & DPO  & 9.5 & 17.9 & 25.5 & 9.5 & \underline{24.7} & 43.2 & \textbf{54.0} & \underline{59.7} & \underline{63.5} & \textbf{54.0} & \underline{64.0} & \underline{73.4}  \\
 & SDPO & 9.7 & \underline{18.0} & 25.2 & 9.7 & \underline{24.7} & 42.4 & 48.1 & 54.3 & 59.0 & 48.1 & 59.0 & 70.4         \\
 & IRPO & \textbf{10.4} & \textbf{19.5} & \textbf{26.8} & \textbf{10.4} & \textbf{25.7} & \underline{43.8} & \underline{53.1} & \textbf{70.1} & \textbf{74.1} & \underline{53.1} & \textbf{82.2} & \textbf{92.0}          \\
\midrule
\multirow{4}{*}{Gemma2} 
 & Base & \underline{27.4} & 35.6 & 42.1 & \underline{27.4} & 42.6 & \underline{58.4} & 27.2 & 35.4 & 41.4 & 27.2 & 42.0 & 56.7 \\
 & SFT  & 23.6 & 32.9 & 39.3 & 23.6 & 40.2 & 55.7 & 29.5 & 38.0 & 44.1 & 29.5 & 44.7 & 59.7 \\
 & DPO  & 27.0 & 34.4 & 40.7 & 27.0 & 40.5 & 56.1 & 29.4 & 37.9 & 43.7 & 29.4 & 44.7 & 59.1 \\
 & SDPO & \textbf{28.7} & \underline{36.5} & \underline{42.6} & \textbf{28.7} & \underline{43.0} & 58.0 & \underline{57.0} & \underline{62.4} & \underline{66.2} & \underline{57.0} & \underline{66.5} & \underline{75.8}         \\
 & IRPO & 27.0 & \textbf{45.5} & \textbf{53.3} & 27.0 & \textbf{59.8} & \textbf{78.7} & \textbf{66.3} & \textbf{79.2} & \textbf{81.7} & \textbf{66.3} & \textbf{88.3} & \textbf{94.5} \\
\bottomrule
\end{tabular}
\caption{
Performance on the ARC and CommonsenseQA QA datasets, evaluated using NDCG and Recall for the top 1, 3, and 5 predictions out of 10 answer choices.
}
\label{tab:main-qa}
\vspace{-1em}
\end{table}

\section{Analysis}

In this section, we analyze the optimization behavior and performance of IRPO in both online and offline settings.
For on-policy optimization (in~\Cref{sec:online}), we extend IRPO to its online variant \textbf{Iterative IRPO}.
Additionally, we present offline learning curves for IRPO across multiple tasks and various backbone LLMs (\Cref{sec:offline}) in \Cref{fig:metric1_list_items}.
We then conduct an ablation study to investigate the role of relevance and positional weighting in IRPO’s objective (\Cref{sec:abl_weight}), demonstrating the necessity of jointly modeling these factors for effective ranking alignment. Finally, we compare IRPO with recent learning-to-rank approaches, focusing on LiPO~\citep{liu2024lipo} (\Cref{sec:ltr}).

\begin{figure}[H]
    \centering
    \begin{subfigure}[t]{0.48\textwidth}
        \centering
        \includegraphics[width=\linewidth]{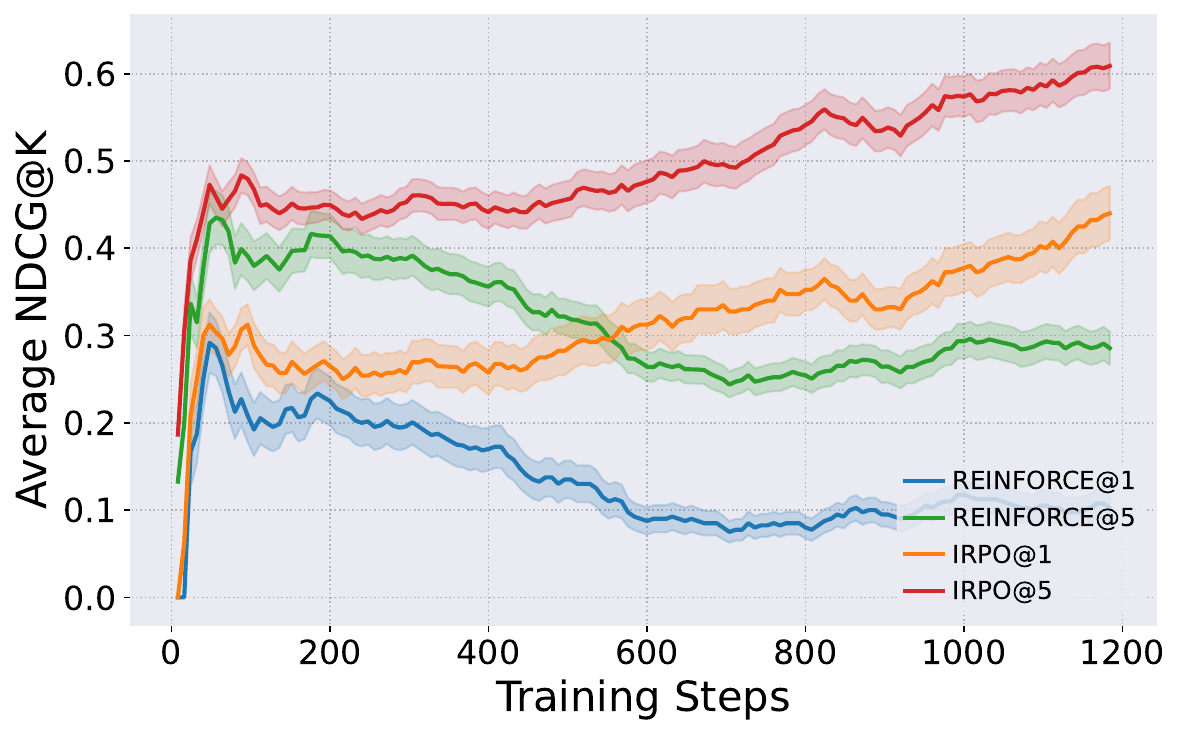}
        \caption{ARC}
        \label{fig:arc_plot}
    \end{subfigure}
    \hfill
    \begin{subfigure}[t]{0.48\textwidth}
        \centering
        \includegraphics[width=\linewidth]{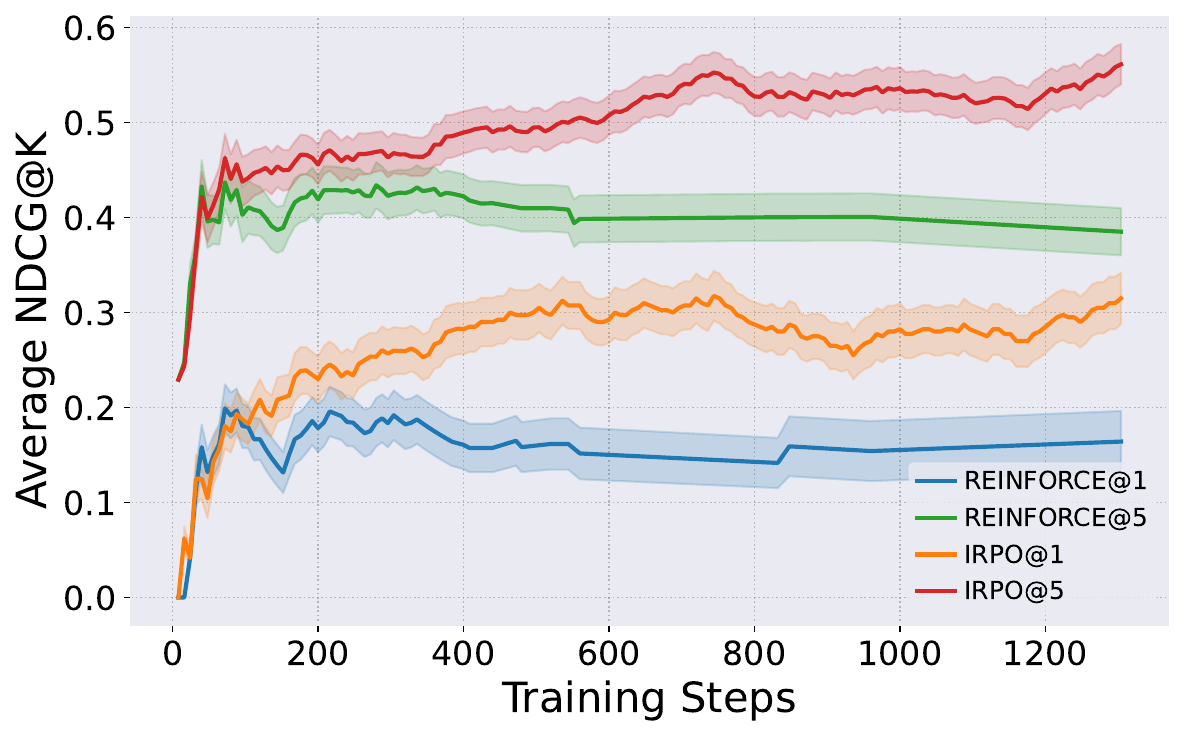}
        \caption{CommonsenseQA}
        \label{fig:commonsenseqa_plot}
    \end{subfigure}
    \caption{Comparison of REINFORCE and IRPO on ARC and CommonsenseQA.}
    \label{fig:ndcg_side_by_side}
\end{figure}

\subsection{On-policy Online Optimization of Iterative IRPO}\label{sec:online}

We explore an on-policy variant of IRPO, \textbf{Iterative IRPO}, which adapts to an online optimization setting \cite{hu2025reinforce++,wu2025ctrls,shao2024deepseekmath}. 
In this setting, models sample their responses based on queries, rather than relying on predefined ranking lists from the original datasets.
These on-policy sampled responses are compared with ground-truth annotations, simulating a realistic human feedback loop.
We conduct such on-policy online learning experiments on ARC and CommonsenseQA, 
compared to a standard policy-gradient baseline, REINFORCE \citep{sutton1999policy}.
In~\Cref{fig:ndcg_side_by_side}, we show that the Iterative IRPO achieves constantly increasing NDCG scores,
while REINFORCE fails to explore effective candidates, leading to insufficient feedback. 
Aligned with the design of IRPO, which prioritizes more relevant items while considering positional importance,
Iterative IRPO could improve the general quality of the entire ranking list,
which significantly benefits on-policy exploration performance.
Supported by our theoretical insights (\Cref{subsec:4.3}), we link IRPO's optimization to an efficient importance sampling method, which inherently serves as an effective exploration mechanism when Iterative IRPO is enabled online. To further illustrate, we provide a qualitative comparison of outputs generated by the base model, Iterative IRPO, and REINFORCE in \Cref{app:case}.

\subsection{Optimization Analysis of IRPO} \label{sec:offline}
\begin{figure}[h]
    \centering
    \includegraphics[width=\textwidth]{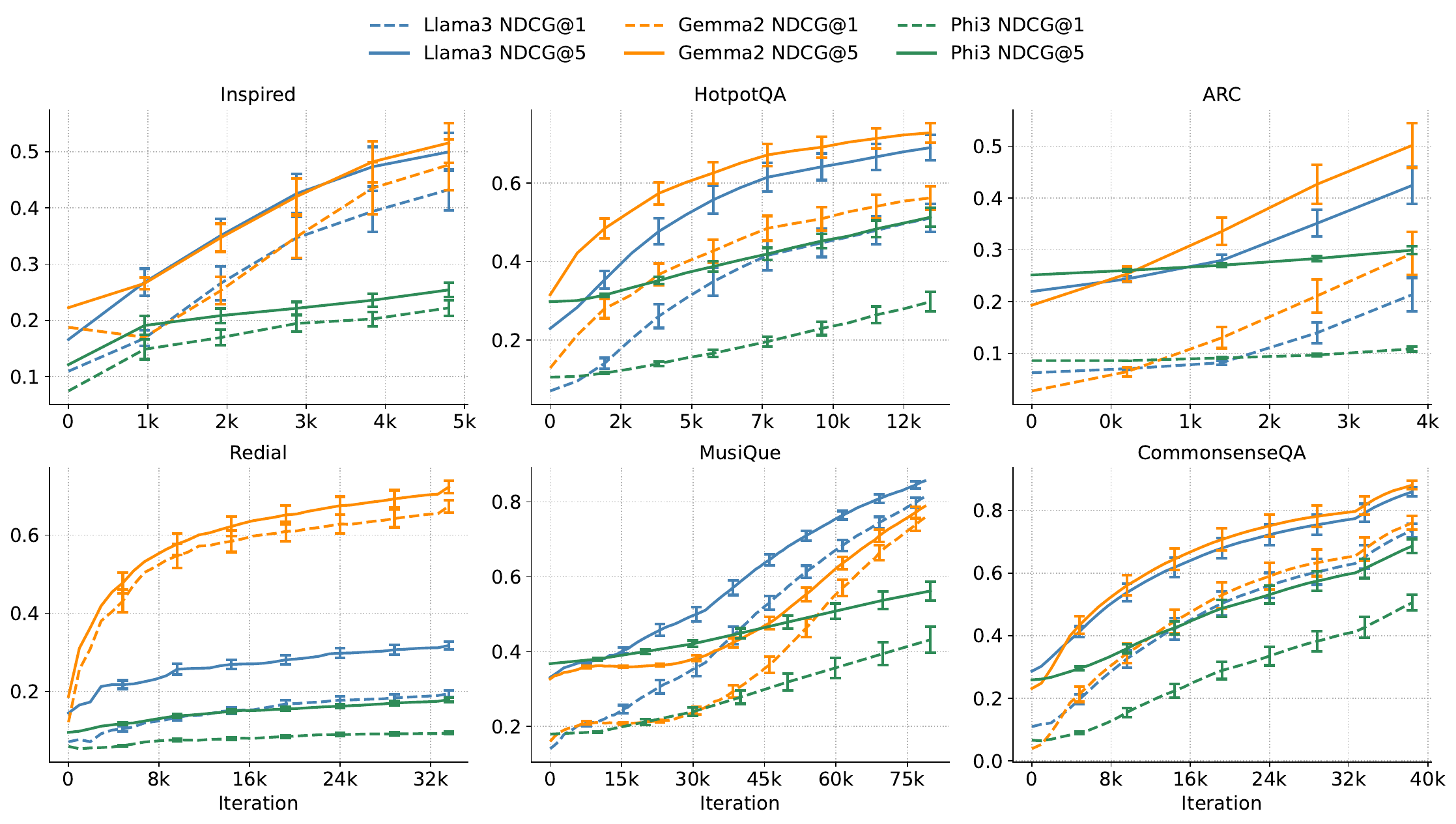}
    \caption{IRPO's performance across six benchmarks using Llama3, Gemma2, and Phi3}
    \label{fig:metric1_list_items}
\end{figure}

We further evaluate IRPO's offline optimization performance 
in the experimental results in \Cref{fig:metric1_list_items},
showing that IRPO consistently achieves higher evaluation NDCG scores and exhibits stable optimization.
across six diverse benchmarks (\textit{Inspired}, \textit{HotpotQA}, \textit{ARC}, \textit{Redial}, \textit{MuSiQue}, and \textit{CommonsenseQA}) using three different LLM backbones: Llama3, Gemma2, and Phi3. 
This robust performance improvement is attributable to IRPO’s adaptive importance weighting mechanism, 
which effectively prioritizes gradient updates toward ranking positions with higher relevance discrepancies.
This mechanism allows IRPO to rapidly capture essential ranking signals from the feedback, leading to stable and consistent optimization behavior,
a finding further supported by our theoretical analysis in~\Cref{subsec:4.3}. 

To highlight the strengths of IRPO over baseline methods, we include three representative qualitative examples in \Cref{app:case_qual_examples}. These showcase how IRPO more effectively ranks contextually relevant and coherent responses.

\subsection{Ablation Study} \label{sec:abl_weight}
IRPO inherently prioritizes relative comparisons over absolute weight magnitudes, making it less sensitive to specific weighting schemes. To validate this, we conduct an additional ablation study evaluating two alternative positional weighting methods: (1) \textbf{abl1:} $w(i) = \frac{1}{\log(1 + i)}$ (positional weights without relevance) (2) \textbf{abl2:} $w(i) = \frac{2^{y_i} - 1}{i}$ (alternative relevance scaling).

As shown in Appendix~\ref{app:ablation_tables}, removing the relevance component (as in \textit{abl1}) leads to significant degradation in ranking performance across all datasets. These results underscore the importance of modeling both item relevance and positional importance within IRPO’s objective.

\subsection{Comparison with Learning-to-Rank (LiPO)}
\label{sec:ltr}
Although IRPO and LiPO~\citep{liu2024lipo} address distinct feedback settings, we provide a comparative analysis here to clearly contextualize IRPO within recent learning-to-rank (LTR) paradigms. LiPO~\citep{liu2024lipo} is a recent representative baseline explicitly extending direct preference optimization (DPO) by integrating general learning-to-rank principles. Unlike IRPO, LiPO is designed primarily for scenarios with fully supervised or extensively labeled listwise data, allowing straightforward integration into standard supervised ranking setups.

To meaningfully compare these distinct approaches, we adapt LiPO to align closely with our sparse, in-context feedback scenario and conduct comparative experiments on representative benchmarks (ARC and MuSiQue). Results summarized in~\Cref{app:lipo_comparison} indicate that IRPO consistently outperforms LiPO, underscoring the advantage of IRPO's explicitly modeled positional relevance and sparse feedback setting.

\section{Conclusion}

In this work, we introduced IRPO, a novel alignment framework that directly optimizes LLMs for ranking tasks using sparse, in-context user feedback. By explicitly modeling both item relevance and positional importance within a differentiable ranking objective, IRPO effectively addresses the limitations of existing DPO methods. Our theoretical insights demonstrated IRPO's adaptive prioritization mechanism and established its connection to importance sampling, providing unbiased gradient estimation with reduced variance. Extensive empirical evaluations across conversational recommendation, generative retrieval, and question-answering re-ranking tasks consistently showed IRPO’s superior ranking performance. Our findings highlight IRPO as an effective and efficient method for aligning LLMs with realistic user preferences, paving the way for broader integration into in-context action-space exploration and reinforcement learning in dynamic online settings.

\bibliographystyle{colm2025_conference}

\appendix

\section{Experimental Details} \label{app:details}
\subsection{Implementation Details}
We validate the effectiveness of IRPO against baseline methods using three popular pre-trained LLMs: 
Llama3.2-3B-Instruct~\citep{grattafiori2024llama}, a 3B-parameter model pretrained on 9 trillion tokens; 
Gemma2-2B-it~\citep{team2024gemma}, a 2B-parameter model pre-trained on 2 trillion tokens; 
and Phi-3-mini-4k-instruct~\citep{abdin2024phi}, 
a 3.8 B-parameter model pre-trained on synthetic and publicly available data, featuring a 4K-token context length.

We used DPO's codebase for implementing both our SFT and DPO baseline experiments. 
For S-DPO, we used the original codebase. 
For IRPO, we implemented our experiments using PyTorch and trained all models using NVIDIA A6000 GPUs.
We set the KL penalty coefficient $\beta$ to 1.0.

\subsection{Run-time Comparison}

IRPO directly aligns LLMs to generate an entire ranked list in a single forward pass, whereas the other baselines require multiple forward passes to calculate pairwise margins. Consequently, IRPO significantly reduces inference time. To illustrate this advantage clearly, we report the average per-sample runtime measured under consistent evaluation settings.~\Cref{tab:runtime_comparison} summarizes these results.

\begin{table}[htbp]
\centering
\begin{tabular}{lc}
\toprule
\textbf{Method} & \textbf{Run Time (per sample)} \\
\midrule
Base  & 0.1140 s \\
DPO   & 0.1145 s \\
SDPO  & 0.1144 s \\
LiPO  & 0.1154 s \\
IRPO  & \textbf{0.0313 s} \\
\bottomrule
\end{tabular}
\caption{Run-time comparison of IRPO and baselines (average per-sample runtime).}
\label{tab:runtime_comparison}
\end{table}

Due to its single-pass architecture, IRPO achieves approximately a $4\times$ speedup in runtime per sample compared to other approaches.

\subsection{Evaluation Details}

\paragraph{Conversational Recommendation}
Inspired~\citep{hayati-etal-2020-inspired}, containing 1,028 dialogues split into 730 training and 88 evaluation samples, 
and Redial~\citep{10.5555/3327546.3327641}, which consists of 10,264 dialogues, 
divided into 8,945 training and 1,319 evaluation samples. 
\textbf{3 ground truth} movies from logged human feedback ranked by popularity in the individual dataset; 
\textbf{5 GPT-generated movies} produced by GPT-3.5 given the same context; 
and \textbf{12 randomly sampled movies} sampled from a frequency-based distribution derived from the training corpus. 

\paragraph{Generative Retrieval}
HotpotQA contains approximately 23K challenging multi-hop questions, split into 15,661 for training and 7,405 for evaluation,
while MuSiQue provides 22,355 questions, divided into 19,938 training and 2,417 evaluation samples.
Each question is associated with supporting sentences labeled as relevant contexts and an additional set of 8 randomly sampled distractor sentences from the same document collection. 
This typically results in 10–12 candidate paragraphs per query.
To ensure context fits within the LLM's context window, we truncate each supporting paragraph to 50 tokens.
Each question is associated with supporting sentences labeled as relevant contexts and an additional set of 8 randomly sampled distractor sentences from the same document collection. 

\paragraph{Question-answering as Re-ranking}
ARC~\citep{clark2018thinksolvedquestionanswering}, comprising 1,418 challenging science-based reasoning questions (1,119 for training and 299 for evaluation), and CommonsenseQA~\citep{talmor-etal-2019-commonsenseqa}, consisting of 10,962 commonsense reasoning questions (9,741 training and 1,221 evaluation). Originally, ARC presents four answer choices per question, while CommonsenseQA includes five.
Specifically, we augment ARC questions with six additional distractors and CommonsenseQA questions with five.
To evaluate LLMs re-ranking capabilities, 
we augment each dataset by introducing additional semantically similar but incorrect answers, 
increasing task complexity, resulting in a uniform set of $10$ candidates per question.

\section{Extending to Other Ranking Metrics}\label{app:metrics}

Our formulation naturally supports many other ranking metrics, such as P@K, MRR, MAP, and eDCG (from easiest to hardest in terms of optimization).
P@K and MRR are simpler compared to the others due to their binary reciprocal structure, making them easier to optimize.
MAP is more complex as it requires normalization, making optimization more difficult.
NDCG and eDCG are the most complex as they have non-linear gradients due to their position-based and relevance-based weighting, leading to more complex updates.

Precision@k (P@k) is binary defined based on whether the item at rank $i$ is relevant within the top $k$ positions:
\[
w(i) = \mathbb{I}(y_{\tau(i)} \geq 1)
, \quad \text{for} \ i \in \{1, 2, \dots, k\},
\]
where $\mathbb{I}(y_{\tau(i)} \geq 1)$ is an indicator function returning $1$ if relevant, and 0 otherwise.

Mean Average Precision (MAP) is the precision at each rank normalized by the total relevant items:
\[
w(i) = \frac{2^{y_{\tau(i)}} - 1}{\sum_{i=1}^{n} \mathbb{I}(y_{\tau(i)} \geq 1)}
\]

Mean Reciprocal Rank (MRR) is based on the reciprocal rank of the first relevant item
\[
w(i) = \frac{1}{\max(r_{\tau(i)}, 1)}, \quad \text{for} \ y_{\tau(i)} \geq 1
\]

The Exponential Discounted Cumulative Gain (eDCG) combines both relevance of an item and an exponential positional discount:
\[
w(i) = \frac{2^{y_{\tau(i)}} - 1}{\exp(\lambda \cdot i)}
\]
where $\lambda$ controls exponential decay with regards to rank.

\section{Derivation of the gradient of IRPO}\label{sec:gradient}

In this section, we provide a detailed derivation of the gradient of the IRPO objective with respect to the model parameters $\theta$. Starting from the IRPO objective in Equation~\eqref{eq:final_obj}, we compute the gradient that forms the basis for our optimization process.

\subsection{IRPO Objective}

Recall that our IRPO objective is defined as:
\begin{align*}
\mathcal{L}_{\text{R-DPO}}(\pi_\theta; \pi_{\text{ref}}) &= -\mathbb{E}_{(x,\boldsymbol{y})}\Biggl[\sum_{i=1}^{n} w(i) \cdot \log \sigma (z_i) \Biggr] \\
w(i) &= \frac{2^{y_{\tau(i)}}-1}{\log_2(1+i)} \\
z_i &= -\log \sum_{j=1}^{n} \exp\!\Bigl(\beta\Bigl[\log \frac{\pi_\theta(e_j\mid x)}{\pi_{\text{ref}}(e_j\mid x)} - \log \frac{\pi_\theta(e_{\tau(i)}\mid x)}{\pi_{\text{ref}}(e_{\tau(i)}\mid x)}\Bigr]\Bigr)
\end{align*}

\subsection{Computing the Gradient}

We compute the gradient of the IRPO objective with respect to the model parameters $\theta$:
\begin{align*}
\nabla_\theta \mathcal{L}_{\text{R-DPO}}(\pi_\theta; \pi_{\text{ref}}) &= -\mathbb{E}_{(x,\boldsymbol{y})}\Biggl[\sum_{i=1}^{n} w(i) \cdot \nabla_\theta \log \sigma (z_i) \Biggr]
\end{align*}

Using the chain rule and properties of the sigmoid function, we can express the gradient of $\log \sigma(z_i)$ as:
\begin{align*}
\nabla_\theta \log \sigma(z_i) &= \frac{1}{\sigma(z_i)} \cdot \nabla_\theta \sigma(z_i) \\
&= \frac{1}{\sigma(z_i)} \cdot \sigma(z_i) \cdot (1 - \sigma(z_i)) \cdot \nabla_\theta z_i \\
&= (1 - \sigma(z_i)) \cdot \nabla_\theta z_i
\end{align*}

Since $1 - \sigma(z_i) = \sigma(-z_i)$, we have:
\begin{align*}
\nabla_\theta \log \sigma(z_i) &= \sigma(-z_i) \cdot \nabla_\theta z_i
\end{align*}

\subsection{Gradient of \texorpdfstring{$z_i$}{zi}}

Next, we compute the gradient of $z_i$:
\begin{align*}
\nabla_\theta z_i &= \nabla_\theta \left( -\log \sum_{j=1}^{n} \exp\!\Bigl(\beta\Bigl[\log \frac{\pi_\theta(e_j\mid x)}{\pi_{\text{ref}}(e_j\mid x)} - \log \frac{\pi_\theta(e_{\tau(i)}\mid x)}{\pi_{\text{ref}}(e_{\tau(i)}\mid x)}\Bigr]\Bigr) \right) \\
&= -\nabla_\theta \log \left( \sum_{j=1}^{n} \exp\!\Bigl(\beta\Bigl[\log \frac{\pi_\theta(e_j\mid x)}{\pi_{\text{ref}}(e_j\mid x)} - \log \frac{\pi_\theta(e_{\tau(i)}\mid x)}{\pi_{\text{ref}}(e_{\tau(i)}\mid x)}\Bigr]\Bigr) \right)
\end{align*}

Let us define:
\begin{align*}
\delta_{ij} &= \log \frac{\pi_\theta(e_j\mid x)}{\pi_{\text{ref}}(e_j\mid x)} - \log \frac{\pi_\theta(e_{\tau(i)}\mid x)}{\pi_{\text{ref}}(e_{\tau(i)}\mid x)} \\
S_i &= \sum_{j=1}^{n} \exp(\beta \delta_{ij})
\end{align*}

Therefore, $z_i = -\log S_i$ and:
\begin{align*}
\nabla_\theta z_i &= -\nabla_\theta \log S_i \\
&= -\frac{1}{S_i} \nabla_\theta S_i \\
&= -\frac{1}{S_i} \nabla_\theta \left( \sum_{j=1}^{n} \exp(\beta \delta_{ij}) \right) \\
&= -\frac{1}{S_i} \sum_{j=1}^{n} \nabla_\theta \exp(\beta \delta_{ij}) \\
&= -\frac{1}{S_i} \sum_{j=1}^{n} \exp(\beta \delta_{ij}) \cdot \beta \cdot \nabla_\theta \delta_{ij}
\end{align*}

\subsection{Gradient of \texorpdfstring{$\delta_{ij}$}{Gradient of delta\_ij}}
Now we compute the gradient of $\delta_{ij}$:
\begin{align*}
\nabla_\theta \delta_{ij} &= \nabla_\theta \left( \log \frac{\pi_\theta(e_j\mid x)}{\pi_{\text{ref}}(e_j\mid x)} - \log \frac{\pi_\theta(e_{\tau(i)}\mid x)}{\pi_{\text{ref}}(e_{\tau(i)}\mid x)} \right) \\
&= \nabla_\theta \log \pi_\theta(e_j\mid x) - \nabla_\theta \log \pi_{\text{ref}}(e_j\mid x) - \nabla_\theta \log \pi_\theta(e_{\tau(i)}\mid x) + \nabla_\theta \log \pi_{\text{ref}}(e_{\tau(i)}\mid x) \\
\end{align*}

Since $\pi_{\text{ref}}$ does not depend on $\theta$, $\nabla_\theta \log \pi_{\text{ref}}(e_j\mid x) = 0$ and $\nabla_\theta \log \pi_{\text{ref}}(e_{\tau(i)}\mid x) = 0$. Therefore:
\begin{align*}
\nabla_\theta \delta_{ij} &= \nabla_\theta \log \pi_\theta(e_j\mid x) - \nabla_\theta \log \pi_\theta(e_{\tau(i)}\mid x) \\
&= \nabla_\theta \log \frac{\pi_\theta(e_j\mid x)}{\pi_\theta(e_{\tau(i)}\mid x)}
\end{align*}

\subsection{Importance Weights and Final Gradient}

Substituting the gradient of $\delta_{ij}$ back into the gradient of $z_i$:
\begin{align*}
\nabla_\theta z_i &= -\frac{1}{S_i} \sum_{j=1}^{n} \exp(\beta \delta_{ij}) \cdot \beta \cdot \nabla_\theta \log \frac{\pi_\theta(e_j\mid x)}{\pi_\theta(e_{\tau(i)}\mid x)} \\
\end{align*}

Defining the importance weights $\rho_{ij}$:
\begin{align*}
\rho_{ij} &= \frac{\exp(\beta \delta_{ij})}{S_i} \\
&= \frac{\exp\!\Bigl(\beta\Bigl[\log \frac{\pi_\theta(e_j\mid x)}{\pi_{\text{ref}}(e_j\mid x)} - \log \frac{\pi_\theta(e_{\tau(i)}\mid x)}{\pi_{\text{ref}}(e_{\tau(i)}\mid x)}\Bigr]\Bigr)}{\sum_{k=1}^{n} \exp\!\Bigl(\beta\Bigl[\log \frac{\pi_\theta(e_k\mid x)}{\pi_{\text{ref}}(e_k\mid x)} - \log \frac{\pi_\theta(e_{\tau(i)}\mid x)}{\pi_{\text{ref}}(e_{\tau(i)}\mid x)}\Bigr]\Bigr)}
\end{align*}

We can now express the gradient of $z_i$ as:
\begin{align*}
\nabla_\theta z_i &= -\beta \sum_{j=1}^{n} \rho_{ij} \cdot \nabla_\theta \log \frac{\pi_\theta(e_j\mid x)}{\pi_\theta(e_{\tau(i)}\mid x)}
\end{align*}

Substituting this into the gradient of $\log \sigma(z_i)$:
\begin{align*}
\nabla_\theta \log \sigma(z_i) &= \sigma(-z_i) \cdot (-\beta) \sum_{j=1}^{n} \rho_{ij} \cdot \nabla_\theta \log \frac{\pi_\theta(e_j\mid x)}{\pi_\theta(e_{\tau(i)}\mid x)} \\
&= -\beta \sigma(-z_i) \sum_{j=1}^{n} \rho_{ij} \cdot \nabla_\theta \log \frac{\pi_\theta(e_j\mid x)}{\pi_\theta(e_{\tau(i)}\mid x)}
\end{align*}

Finally, substituting into the gradient of the IRPO objective:
\begin{align*}
\nabla_\theta \mathcal{L}_{\text{R-DPO}}(\pi_\theta; \pi_{\text{ref}}) &= -\mathbb{E}_{(x,\boldsymbol{y})}\Biggl[\sum_{i=1}^{n} w(i) \cdot \nabla_\theta \log \sigma (z_i) \Biggr] \\
&= -\mathbb{E}_{(x,\boldsymbol{y})}\Biggl[\sum_{i=1}^{n} w(i) \cdot \Bigl(-\beta \sigma(-z_i) \sum_{j=1}^{n} \rho_{ij} \cdot \nabla_\theta \log \frac{\pi_\theta(e_j\mid x)}{\pi_\theta(e_{\tau(i)}\mid x)}\Bigr) \Biggr] \\
&= \beta\mathbb{E}_{(x,\boldsymbol{y})}\Biggl[\sum_{i=1}^{n} w(i) \sigma(-z_i) \sum_{j=1}^{n} \rho_{ij} \cdot \nabla_\theta \log \frac{\pi_\theta(e_j\mid x)}{\pi_\theta(e_{\tau(i)}\mid x)} \Biggr]
\end{align*}

\section{Mean Analysis} \label{app:mean}
\paragraph{Lemma 4.1} Let $\hat{g}(e_j)$ be defined as in \eqref{eq:gd-estimator}. Each $e_j$ is sampled from $e$ with the probability of $\rho_{i,j}=p(e_j)$, at position $i$ in the ranking list. Then $\mathbb{E}\left[\hat{g}(e_j)\right]=g$ in \eqref{eq:gd}.

\textit{Proof.}~ Following \citep{}, we derive the proof by a sequence of identities.
\begin{align*}
    \mathbb{E}\Bigl[\hat{g}(e_j)\Bigr] 
    &=  \mathbb{E}_{e_j \sim \boldsymbol{e}}\Bigl[\hat{g}(e_j) \Bigr] \\
    &=  \mathbb{E}_{e_j \sim \boldsymbol{e}}\Bigl[\nabla_\theta \log \frac{\pi_\theta(e_j\mid x)}{\pi_\theta(e_{\tau(i)}\mid x)} \Bigr] \\
    &=  \sum_{j=1}^n \rho_{ij} \cdot \nabla_\theta \log \frac{\pi_\theta(e_j\mid x)}{\pi_\theta(e_{\tau(i)}\mid x)} = g.
\end{align*}

\section{Variance Analysis} \label{app:var}
\paragraph{Lemma 4.2} 
With the assumption of $L=\max_{j}{[\nabla_\theta \log \pi_\theta(e_j\mid x) - \nabla_\theta \log \pi_\theta(e_{\tau(i)}\mid x)]}$,
we derive the expected absolute deviation of the gradient estimation $\mathbb{E}[|\hat{g}(e_j) - g|] $.

\textit{Proof.}~ By Cauchy-Schwartz inequality \citep{}, we have
\begin{equation*}
    \mathbb{E}[|\hat{g}(e_j)-g|] \le \sqrt{\mathbb{E}\Bigl[(\hat{g}(e_j)-g)^2\Bigr]}.
\end{equation*}
Consider the variance of the estimator, $\operatorname{Var}\bigl(\hat{g}(e_j)\bigr) = \mathbb{E}\Bigl[(\hat{g}(e_j)-g)^2\Bigr]$.
We sample according to the importance weights $p(e_j)=\rho_{ij}$. 
Following standard importance sampling \citep{robert1999monte,shapiro2003monte}, we further obtain the estimation upper bound by a factor proportional to the maximum importance weight,
\begin{equation*}
    \operatorname{Var}\bigl(\hat{g}(e_j)\bigr) \le \frac{\|\rho_{i,j}\|_\infty}{n}\,\mathbb{E}\Bigl[\|z(e_j)\|^2\Bigr],
\end{equation*}
where $z(e_j) = \nabla_\theta \log \pi_\theta(e_j\mid x) - \nabla_\theta \log \pi_\theta(e_{\tau(i)}\mid x)$, and the variance is reduced by the factor of $n$ independent samples.
Substituting the variance bound into the expected absolute deviation, we have
\begin{equation*}
    \mathbb{E}[|\hat{g}(e_j)-g|] \le \sqrt{\frac{\|\rho_{i,j}\|_\infty}{n}\,\mathbb{E}\Bigl[\|z(e_j)\|^2\Bigr]}.
\end{equation*}
With the assumption of bounded gradient difference $L=\max_{j}{[\nabla_\theta \log \pi_\theta(e_j\mid x) - \nabla_\theta \log \pi_\theta(e_{\tau(i)}\mid x)]}$, 
which in practice is achieved by gradient clipping \citep{ouyang2022training, schulman2017proximal}, we achieve the final bound as 
\begin{equation*}
    \mathbb{E}[|\hat{g}(e_j)-g|] \le \sqrt{\frac{\|\rho_{i,j}\|_\infty}{n}\,L^2} = L\sqrt{\frac{\|\rho_{i,j}\|_\infty}{n}}.
\end{equation*}

\section{Additional Results}
\label{app:ablation_tables}

\subsection{Ablation Study Results}

We compare these variants using the Llama3 backbone across three benchmarks, representing each task category: Inspired (conversational recommendation), MusiQue (generative retrieval), and ARC (question-answering re-ranking). 

We provide detailed ablation results in ~\Cref{tab:abl_inspired}--\ref{tab:abl_arc}, evaluating alternative positional weighting schemes across three benchmark tasks.

\begin{table}[hbtp]
\centering
\begin{tabular}{lcccccc}
\toprule
\textbf{Method} & N@1 & N@5 & N@10 & R@1 & R@5 & R@10 \\
\midrule
IRPO         & 45.3 & 55.2 & 62.3 & 22.4 & 68.8 & 87.2 \\
IRPO (abl1)  & 35.0 & 37.9 & 48.5 & 18.9 & 46.0 & 72.9 \\
IRPO (abl2)  & 43.8 & 49.2 & 58.8 & 21.0 & 58.9 & 83.1 \\
\bottomrule
\end{tabular}
\caption{Ablation results on the Inspired dataset.}
\label{tab:abl_inspired}
\end{table}

\begin{table}[hbtp]
\centering
\begin{tabular}{lcccccc}
\toprule
\textbf{Method} & N@1 & N@3 & N@5 & R@1 & R@3 & R@5 \\
\midrule
IRPO         & 65.1 & 59.4 & 69.4 & 27.6 & 59.2 & 78.0 \\
IRPO (abl1)  & 50.9 & 54.0 & 65.8 & 21.0 & 56.9 & 78.9 \\
IRPO (abl2)  & 68.3 & 62.8 & 72.7 & 29.0 & 62.9 & 81.5 \\
\bottomrule
\end{tabular}
\caption{Ablation results on the MuSiQue dataset.}
\label{tab:abl_musique}
\end{table}

\begin{table}[bhtp]
\centering
\begin{tabular}{lcccccc}
\toprule
\textbf{Method} & N@1 & N@3 & N@5 & R@1 & R@3 & R@5 \\
\midrule
IRPO         & 27.4 & 38.9 & 46.5 & 27.4 & 47.6 & 66.3 \\
IRPO (abl1)  &  9.4 & 17.2 & 23.4 &  9.4 & 22.9 & 38.2 \\
IRPO (abl2)  & 27.1 & 37.8 & 46.1 & 27.1 & 45.8 & 66.3 \\
\bottomrule
\end{tabular}
\caption{Ablation results on the ARC dataset.}
\label{tab:abl_arc}
\end{table}

\subsection{Comparision with LiPO (LTR)}
\label{app:lipo_comparison}
We report detailed comparisons between IRPO and LiPO on the ARC and MuSiQue datasets. Tables~\ref{tab:arc_lipo_comparison} and~\ref{tab:musique_lipo_comparison} present results for both LLaMA3 and Gemma2 backbones, demonstrating IRPO's consistent advantage.

\begin{table}[htbp]
\centering
\begin{tabular}{llcccccc}
\toprule
\textbf{Model} & \textbf{Method} & N@1 & N@3 & N@5 & R@1 & R@3 & R@5 \\
\midrule
\multirow{5}{*}{LLaMA3}
& Base  & 13.2 & 21.7 & 28.6 & 13.2 & 28.4 & 44.9 \\
& SFT   & 13.5 & 21.4 & 28.4 & 13.5 & 27.7 & 44.6 \\
& SDPO  & 23.4 & 28.3 & 35.4 & 23.4 & 32.5 & 50.0 \\
& LiPO  & 35.2 & 39.8 & 45.5 & 35.2 & 43.7 & 57.7 \\
& IRPO  & 27.4 & 38.9 & 46.5 & 27.4 & 47.6 & 66.3 \\
\midrule
\multirow{5}{*}{Gemma2}
& Base  & 27.4 & 35.6 & 42.1 & 27.4 & 42.6 & 58.4 \\
& SFT   & 23.6 & 32.9 & 39.3 & 23.6 & 40.2 & 55.7 \\
& SDPO  & 28.7 & 36.5 & 42.6 & 28.7 & 43.0 & 58.0 \\
& LiPO  & 23.2 & 30.5 & 37.6 & 23.2 & 36.0 & 53.5 \\
& IRPO  & 27.0 & 45.5 & 53.3 & 27.0 & 59.8 & 78.7 \\
\bottomrule
\end{tabular}
\caption{Performance comparison on ARC.}
\label{tab:arc_lipo_comparison}
\end{table}

\begin{table}[htbp]
\centering
\begin{tabular}{llcccccc}
\toprule
\textbf{Model} & \textbf{Method} & N@1 & N@3 & N@5 & R@1 & R@3 & R@5 \\
\midrule
\multirow{5}{*}{LLaMA3}
& Base  & 42.5 & 43.5 & 51.9 & 18.8 & 44.9 & 60.7 \\
& SFT   & 36.9 & 39.1 & 48.0 & 16.6 & 40.7 & 57.6 \\
& SDPO  & 58.6 & 60.8 & 66.7 & 26.0 & 62.3 & 73.3 \\
& LiPO  & 47.4 & 49.6 & 57.1 & 20.9 & 51.1 & 65.2 \\
& IRPO  & 65.1 & 59.4 & 69.4 & 27.6 & 59.2 & 78.0 \\
\midrule
\multirow{5}{*}{Gemma2}
& Base  & 40.9 & 42.8 & 51.1 & 17.9 & 44.3 & 59.9 \\
& SFT   & 39.8 & 43.3 & 51.7 & 17.2 & 45.3 & 61.3 \\
& SDPO  & 40.6 & 42.3 & 50.6 & 17.9 & 43.7 & 59.4 \\
& LiPO  & 49.8 & 51.8 & 58.9 & 22.3 & 53.2 & 66.7 \\
& IRPO  & 55.4 & 51.0 & 61.1 & 23.8 & 51.6 & 70.6 \\
\bottomrule
\end{tabular}
\caption{Performance comparison on MuSiQue.}
\label{tab:musique_lipo_comparison}
\end{table}

\section{Case Study} 
\subsection{On-policy} \label{app:case}
\begin{answerbox}{Q: Pretend you are a question-answering model. I will give you a Question. Based on the Question please rerank the answers in a list from the best to the worst without any extract words. Question: At what level of organization does damage compromise the function of the whole system? (A) ecosystems can be damaged \textbf{\textcolor{correctgreen}{(B) an organ}} (C) a cell (D) alters the behavior of the organisms (E) the structure of the organism's anatomy (F) in response to an altering environment. (G) their views of storm damages (H) different systems working together (I) an organelle (J) a tissue
}
\textbf{Base Llama3:} \\
Here is the list of answers in order from best to worst: \textbf{\textcolor{wrongred}{1. a cell}} 2. an organ 3. a tissue 4. an organelle 5. the structure of the organism\'s anatomy 6. different systems working together 7. ecosystems can be damaged 8. in response to an altering environment 9. alters the behavior of the organisms 10. their views of storm damages

\vspace{1em}
\textbf{IRPO Llama-3:} \\
Here is the reranked list of answers from best to worst: \textbf{\textcolor{correctgreen}{1. an organ}} 2. a tissue 3. an organelle 4. a cell 5. different systems working together. 6. the structure of the organism\'s anatomy 7. alters the behavior of the organisms. 8. their views of storm damages 9. in response to an altering environment. 10. ecosystems can be damaged.

\vspace{1em}

\textbf{REINFORCE Llama-3:} \\
Based on the question, here is the reranked list from best to worst answer: \textbf{\textcolor{wrongred}{1. Different systems working together. (This answer implies a high level of organization and interconnectedness, which makes it more likely to compromise the function of the whole system.)}} 2. A tissue (A tissue is a group of cells working together, so it\'s a plausible answer that could compromise the function of the whole system.) 3. An organ (An organ is a group of tissues working together, so it\'s also a plausible answer that could compromise the function of the whole system.) 4. A cell (A cell is the basic unit of organization, and damage to a cell could potentially compromise the function of the whole system.) 5. The structure of the organism\'s anatomy (This answer is more specific to the internal structure of an organism, which may not necessarily compromise the function of the whole system.) 6. Ecosystems can be damaged (This answer is more general and doesn't necessarily imply a level of organization that would compromise the function of the whole system.) 7. Alters the behavior of the organisms. (This answer is more related to the impact of damage on the organisms themselves, rather than the organization of the system.) 8. In response to an altering environment. (This answer is more related to the response of the system to environmental changes, rather than the organization of the system itself.) 9. Their views of storm damages (This answer is unrelated to the organization of the system and more related to opinions or perspectives.) 10. An organelle (An organelle is a specialized structure within a cell, and damage to an organelle might not necessarily compromise the function of the whole system.) Note that this ranking is subjective and based on my understanding of the question. The best answer may depend on the specific context and requirements of the question.

\end{answerbox}

\vspace{1em}

\begin{answerbox}{Q: Pretend you are a question-answering model. I will give you a Question. Based on the Question please rerank the answers in a list from the best to the worst without any extract words. Question: What happens while chatting with friends about current events? (A) laughter (B) hope (C) social event (D) feel close to (E) talking with others (F) meeting friends (G) event (H) talk with friends \textbf{\textcolor{correctgreen}{(I) exchanging information}} (J) social approval}

\textbf{Base Llama3:} \\
Here is the reranked list of answers from best to worst based on their relevance and usefulness in responding to the question: \textbf{\textcolor{wrongred}{1. talking with others}} 2. exchanging information 3. social event 4. social approval 5. laugh 6. feel close to 7. meeting friends 8. event 9. talk with friends 10. hope

\textbf{IRPO Llama-3:} \\
Based on the question ``What happens while chatting with friends about current events?", here is the list of answers from best to worst without any extraneous words: \textbf{\textcolor{correctgreen}{1. exchanging information}} 2. social approval 3. talking with others 4. social event 5. event 6. laugh 7. hope 8. talk with friends 9. meet friends 10. feel close to
\vspace{1em}

\textbf{REINFORCE Llama-3:} \\
 Based on the question, here is the reranked list of answers from best to worst: \textbf{\textcolor{wrongred}{1. talking with others}} 2. talking with friends 3. talk with friends 4. social event 5. exchanging information 6. laugh 7. hope 8. feel close to 9. meeting friends 10. event
\end{answerbox}

\vspace{1em}

\begin{answerbox}{Q: Pretend you are a question-answering model. I will give you a Question. Based on the Question please rerank the answers in a list from the best to the worst without any extract words. Question: Why does a person stay in bed on the weekend? (A) get into bed (B) bedsores (C) slumber (D) lay down (E) get in bed (F) to sleep (G) rest in bed (H) falling asleep \textbf{\textcolor{correctgreen}{(I) sleep in}} (J) stay in bed}
\textbf{Base Llama3:} \\
Based on the context of the question, here is the reranked list from best to worst: \textbf{\textcolor{wrongred}{1. stay in bed}} 2. sleep in 3. get into bed 4. lay down 5. rest in bed 6. falling asleep 7. to sleep 8. get in bed 9. bedsores (this one is out of context and unrelated to the question) 10. slumber (this one is a verb that means to sleep, but it's not as directly relevant to the question as the other options)
\vspace{1em}

\textbf{IRPO Llama-3:} \\
Here is the list of answers in order from best to worst, without any extraneous words: \textbf{\textcolor{correctgreen}{1. sleep in}} 2. rest in bed 3. lay down 4. get into bed 5. get in bed 6. stay in bed 7. to sleep 8. slumber 9. falling asleep 10. bedsores
\vspace{1em}

\textbf{REINFORCE Llama-3:} \\
Based on the Question, here is the reranked list of answers from best to worst: 1. \textbf{\textcolor{wrongred}{stay in bed (Most relevant and accurate answer)}} 2. sleep in (Very relevant and accurate answer) 3. lay down (Relevant answer, but not as specific as "stay in bed") 4. rest in bed (Relevant answer, but not as specific as "stay in bed") 5. get into bed (Relevant answer, but not as specific as "stay in bed") 6. get in bed (Similar to "get into bed", but with a slight variation in wording) 7. to sleep (Not as relevant or specific as other answers) 8. slumber (Not directly related to the question) 9. bedsores (Not relevant to the question at all) 10. falling asleep (Not a complete answer, as it\'s a state rather than an action) Please note that the ranking may vary depending on the context and the users intent.
\end{answerbox}

\subsection{Qualitative examples}
\label{app:case_qual_examples}

\begin{answerbox}{Q. Pretend you are a question-answering model. I will give you a Question. Based on the Question please rerank the provided answers in a list from the best to the worst without any extra words.Question: Sam wasn't lying but he left out important details. He was being what? 1. dishonesty \textbf{\textcolor{correctgreen}{2. deceitful}} 3. lying 4. deceptive 5. imagination 6. more lies 7. lie to himself 8. deception 9. telling lies 10. poker face}

\textbf{SFT:} \\
 \textbf{\textcolor{wrongred}{1. dishonesty}} 2. deceitful 3. lying 4. deceptive 5. imagination

\textbf{DPO:} \\
 \textbf{\textcolor{wrongred}{1. deceptive}} 2. lying 3. deceitful 4. deceptive 5. imagination
 
\textbf{IRPO:} \\
\textbf{\textcolor{correctgreen}{1. deceitful}} 2. dishonesty 3. more lies 4. deceptive 5. imagination 6. lie to himself 7. deception 8. telling lies 9. poker face 10. lying

\end{answerbox}
\vspace{1em}

\begin{answerbox}{Q. Pretend you are a question-answering model. I will give you a Question. Based on the Question please rerank the provided answers in a list from the best to the worst without any extra words. Question: At the end of the day as he began to eat he paused and thanked her it wasn't often she would what? \textbf{\textcolor{correctgreen}{1. cook dinner}} 2. get food 3. make food 4. did chores 5. stretch out 6. wife asked to 7. when holding her hand 8. a meal 9. were hungry 10. she cannot do that when all are for her}

\textbf{SFT:} \\
 \textbf{\textcolor{wrongred}{1. wife asked to}} 2. when holding her hand 3. she cannot do that when all are for her 4. she would 5. cook dinner 6. get food 7. make food 8. did chores 9. stretch out 10. were hungry

\textbf{DPO:} \\
 \textbf{\textcolor{wrongred}{1. stretch out}} 2. cook dinner 3. get food 4. make food 5. did chores 6. wife asked to 7. when holding her hand 8. were hungry 9. she cannot do that when all are for her
 
\textbf{IRPO:} \\
\textbf{\textcolor{correctgreen}{1. cook dinner}}  2. she cannot do that when all are for her 3. wife asked to 4. get food 5. a meal 6. stretch out 7. when holding her hand 8. make food 9. were hungry 10. did chores

\end{answerbox}

\vspace{1em}

\begin{answerbox}{Q. Pretend you are a question-answering model. I will give you a Question. Based on the Question please rerank the provided answers in a list from the best to the worst without any extra words. Question: No matter what date you put on it we all know the universe to be what? \textbf{\textcolor{correctgreen}{1. very old}} 2. infinite 3. infiniverse 4. real 5. universe 6. getting younger 7. miss universe 8. the universe 9. astronomy 10. date}

\textbf{SFT:} \\
 \textbf{\textcolor{wrongred}{1. infinite}} 2. astronomy 3. the universe 4. real 5. getting younger 6. miss universe 7. date 8. infiniverse 9. very old
 
\textbf{DPO:} \\
 \textbf{\textcolor{wrongred}{1. infinite}} 2. astronomy 3. the universe 4. real 5. getting younger 6. universe 7. miss universe 8. date 9. infiniverse 10. very old
 
\textbf{IRPO:} \\
\textbf{\textcolor{correctgreen}{1. very old}} 2. infinite 3. infiniverse 4. real 5. universe 6. getting younger 7. miss universe 8. the universe 9. astronomy 10. date

\end{answerbox}

\vspace{1em}

\end{document}